\documentclass[lettersize,journal]{IEEEtran}
\usepackage{amsmath,amsfonts}
\usepackage{algorithm}
\usepackage{algpseudocode}
\usepackage{array}
\usepackage[caption=false,font=normalsize,labelfont=sf,textfont=sf]{subfig}
\usepackage{textcomp}
\usepackage{stfloats}
\usepackage{url}
\usepackage{verbatim}
\usepackage{graphicx}
\usepackage{cite}
\usepackage{tabularx}
\usepackage{booktabs}
\usepackage{multirow}
\usepackage{makecell}
\usepackage{xcolor}

\begin{document}

\title{Guided Diffusion-based Generation of Adversarial Objects for Real-World Monocular Depth Estimation Attacks}

\author{Yongtao Chen, Yanbo Wang, Wentao Zhao, Guole Shen, Tianchen Deng, Jingchuan Wang,~\IEEEmembership{Senior Member,~IEEE}

\thanks{Yongtao Chen, Yanbo Wang, Wentao Zhao, Guole Shen, Tianchen Deng and Jingchuan Wang are with the School of Automation and Intelligent Sensing, Institute of Medical Robotics, Shanghai Jiao Tong University, Shanghai 200240, China. Yongtao Chen and Yanbo Wang contributed equally to this work. Jingchuan Wang (jchwang@sjtu.edu.cn) is the corresponding authors.}}%

\markboth{Journal of \LaTeX\ Class Files,~Vol.~14, No.~8, August~2021}%
{Shell \MakeLowercase{\textit{et al.}}:Guided Diffusion-based Generation of Adversarial Objects for Real-World Monocular Depth Estimation Attacks}

\maketitle

\begin{abstract}

Monocular Depth Estimation (MDE) serves as a core perception module in autonomous driving systems, but it remains highly susceptible to adversarial attacks. Errors in depth estimation may propagate through downstream decision making and influence overall traffic safety. Existing physical attacks primarily rely on texture-based patches, which impose strict placement constraints and exhibit limited realism, thereby reducing their effectiveness in complex driving environments. To overcome these limitations, this work introduces a training-free generative adversarial attack framework that generates naturalistic, scene-consistent adversarial objects via a diffusion-based conditional generation process. The framework incorporates a Salient Region Selection module that identifies regions most influential to MDE and a Jacobian Vector Product Guidance mechanism that steers adversarial gradients toward update directions supported by the pre-trained diffusion model. This formulation enables the generation of physically plausible adversarial objects capable of inducing substantial adversarial depth shifts. Extensive digital and physical experiments demonstrate that our method significantly outperforms existing attacks in effectiveness, stealthiness, and physical deployability, underscoring its strong practical implications for autonomous driving safety assessment.
\end{abstract}

\section{Introduction}

\IEEEPARstart{A}{utonomous} driving systems have widely adopted Monocular Depth Estimation (MDE) \cite{monodepth2,depthhints,manydepth,MonoDEVSNet,depthanything,depth5,depth6} to either explicitly perceive road geometry and estimate distances to surrounding objects, or implicitly serve as a geometric feature encoder in the upstream of end-to-end networks. 
MDE refers to the task of predicting dense scene depth from a single RGB image by leveraging visual cues such as perspective, occlusion, and object scale. 
Accurate depth estimation is critical for driving safety, as it underpins essential functions such as road-surface understanding \cite{lane}, collision avoidance \cite{collision}, and motion planning \cite{nav} in complex real-world environments.
Notably, Tesla has already integrated MDE into their production-grade vehicles, making it a core component of their perception stack \cite{teslaAI,teslaHackerVision,karpathyFSD}.

In recent years, Deep Neural Networks (DNNs) have demonstrated remarkable performance in MDE, significantly advancing perception in autonomous driving system \cite{mgnet}. 
However, ensuring the security and robustness of DNN-based perception remains a significant challenge~\cite{advtits}.
These models are inherently sensitive to distribution shifts and can produce erroneous predictions under small perturbations, i.e., adversarial examples~\cite{advexample,digital1,digital2,sradv}. 

In the context of autonomous driving within intelligent transportation systems, misestimated depth can lead to hazardous behaviors, including premature braking, unsafe following distances, and incorrect obstacle avoidance. Importantly, perception induced decision errors are not confined to individual vehicles. In real traffic environments, abnormal driving behaviors caused by depth misestimation may propagate through surrounding traffic, resulting in cascading unsafe maneuvers, widespread braking events, and an overall degradation of traffic safety. Moreover, in vehicle and infrastructure cooperative settings, perception outputs may be shared or jointly utilized across multiple agents. As a result, erroneous depth estimates can influence collective environmental understanding, thereby amplifying their impact at the traffic system level.

While adversarial examples were originally studied as a tool for analyzing and improving model robustness~\cite{robustmde}, recent research has increasingly focused on adversarial behaviors that persist in the physical world~\cite{phmde,advrm}. Physical-world adversarial examples expose vulnerabilities that cannot be captured by digital-domain analyses alone and thus are essential for evaluating the reliability of deployed systems.

In the digital domain, adversarial examples are typically crafted by injecting imperceptibly small perturbations into input images, significantly degrading DNN performance~\cite{digital1,digital2}. However, these attacks assume direct access to digital inputs and ignore real-world imaging effects. In contrast, physical-world adversarial examples must be materialized in the environment and remain effective under variations in illumination, and sensor noise, making them substantially more challenging to design. Despite these challenges, physical attacks pose greater safety risks because they can realistically occur in real-world traffic environments and directly compromise deployed perception systems without requiring access to internal models or communication channels.

Prior work has explored physical adversarial attacks on MDE by designing printable patches that can be attached to scene elements to perturb depth predictions~\cite{phy1,phy2,phy3}. Later efforts improved attack stealthiness by placing adversarial patches directly on objects to manipulate their perceived depth~\cite{phmde}, or on the ground to exploit MDE’s reliance on road geometry~\cite{advrm}.

\begin{figure*}[!t]
\centering
\includegraphics[width=\textwidth]{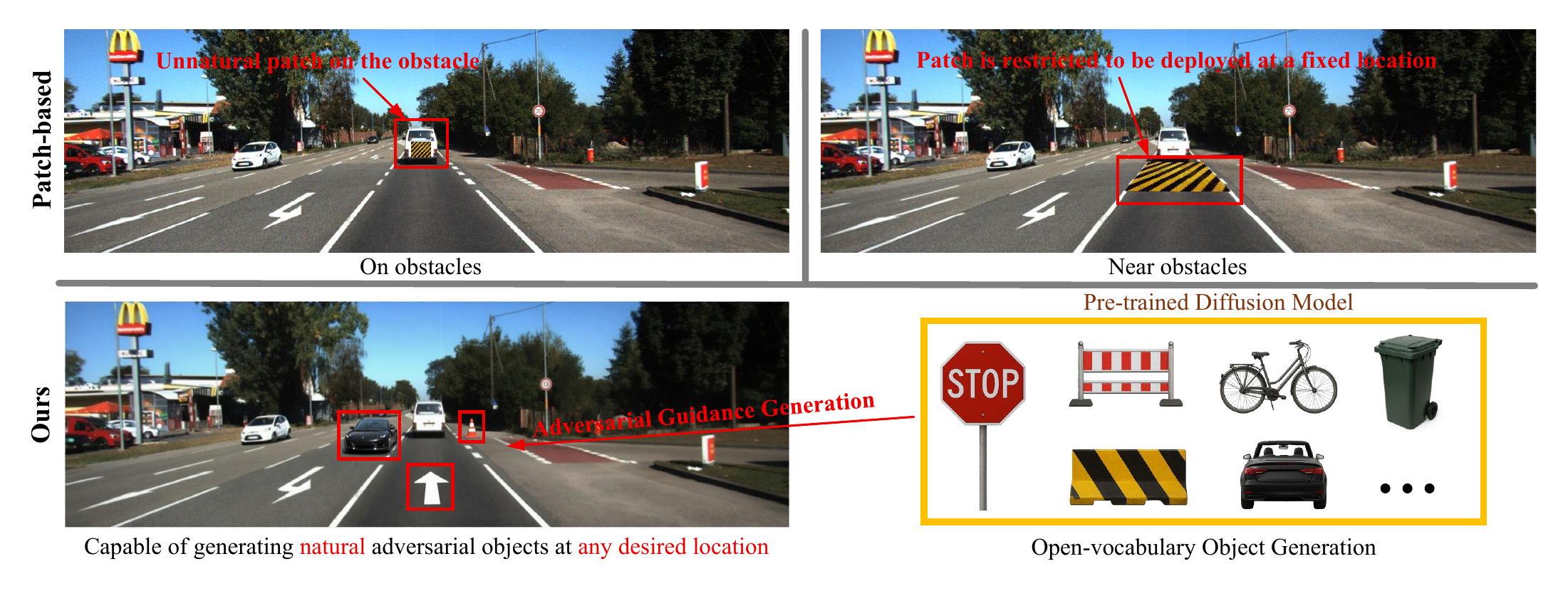}
\caption{Comparison between our generative adversarial object attack and previous patch-based physical attacks. Previous patch-based methods are constrained to fixed spatial locations and rely on unnatural textures, which are prone to being detected and filtered by anomaly or out-of-distribution detectors. In contrast, our method leverages open-vocabulary object generation to generate natural object-level adversarial content that can be flexibly placed at any region.}
\label{compare}
\end{figure*}

Despite these advances, existing physical attacks on MDE predominantly rely on localized texture-based patches deliberately placed in constrained regions for stealth. This patch-based paradigm introduces two practical limitations, as illustrated in Fig.~\ref{compare}. First, patches often exhibit unnatural textures or sharp boundaries that hinder seamless integration into diverse scenes. Second, effective attacks require careful spatial placement, limiting flexibility and reducing practicality in real driving environments. These limitations motivate moving beyond texture patches toward physically realizable adversarial objects that are both naturalistic and semantically coherent.

To this end, we explore an alternative paradigm based on conditional generation\cite{cg,sddiffusion,ppt,brushnet}. Instead of placing artificial patches, this paradigm leverages generative models to generate naturalistic objects that seamlessly blend into the scene. With the introduction of a novel guidance mechanism, the generative process can be effectively steered toward adversarial objectives while preserving the plausibility and visual coherence of the synthesized content.

In contrast to prior attacks on MDE, our method is probably the first training-free approach that moves beyond patch-based strategies by generating natural and semantically meaningful objects suitable for diverse and complex traffic scenarios. Our key observation is that MDE models rely on holistic scene cues for depth estimation, enabling inpainting as a viable attack strategy.
We conceptualize this as a new guidance task balancing two competing objectives: (i) preserving semantic plausibility for natural scene integration and (ii) maximizing adversarial impact on depth predictions. This formulation opens a novel direction for studying controllable generation under safety-critical constraints.

Our main contributions can be summarized as follows:  
\begin{enumerate}
    \item We propose a training-free physical-world adversarial attack framework which formulates the attack as a generative problem, leading to superior attack effectiveness and enhanced stealth.

    \item A novel training-free guidance mechanism is introduced to modulate adversarial updates according to the geometric characteristics implicit in a pre-trained diffusion model, enabling the generation of naturalistic and physically coherent adversarial objects suitable for real-world deployment.

    \item Extensive experiments demonstrate that the proposed method can induce erroneous depth estimates across mainstream MDE models and can be physically realized through printed adversarial objects deployed in real-world environments. Beyond attack effectiveness, the results highlight critical directions for strengthening geometric robustness in vision-based depth perception.
\end{enumerate}

\section{RELATED WORK}

\subsection{Monocular Depth Estimation (MDE)}

MDE is a fundamental perception task that infers 3D scene structure from a single RGB image~\cite{monodepth2,depthhints} and serves as a core component of modern autonomous driving systems. Recent advances in deep learning have substantially improved MDE performance, evolving from early convolutional methods~\cite{monodepth2,depthhints,MonoDEVSNet,manydepth} to transformer-based architectures that leverage global self-attention for enhanced contextual reasoning and generalization~\cite{vit1,depthanything,vit2}. Despite these developments, MDE models remain highly susceptible to adversarial perturbations, where both imperceptible digital modifications and physically realizable perturbations can induce significant depth misestimations. Motivated by these vulnerabilities, our work introduces a diffusion-based adversarial guidance framework that generates realistic, scene-consistent adversarial objects capable of compromising MDE in real-world environments.

\subsection{Physical Adversarial Attack on MDE}

Physical adversarial attacks on MDE have primarily relied on patch-based strategies. Early work optimized printable adversarial patches that could be placed in real environments to perturb depth predictions~\cite{phy1,phy2,phy3}. Subsequent studies improved stealthiness by embedding patches into semantically meaningful regions, for example directly on obstacles~\cite{phmde} or on road surfaces~\cite{advrm}, thereby exploiting the contextual and geometric priors inherent to MDE models. However, patch-based attacks remain fundamentally limited by their reliance on 2D texture patterns, which often introduce unnatural appearance and require placement within restricted spatial regions to remain inconspicuous.
In contrast, our work moves beyond texture-based manipulation and introduces a conditional diffusion framework that generates realistic, semantically coherent adversarial objects, offering greater spatial flexibility, stronger scene integration, and improved physical-world attack effectiveness.

\subsection{Conditional Generation}
Diffusion models~\cite{ddpm,ddim} have recently demonstrated remarkable generative capability across diverse domains~\cite{diffusionsurvey,deng2025}, becoming a dominant paradigm for controllable image generation. 
Conditional generation extends diffusion models by guiding the sampling process toward a desired condition \( c \), typically by modifying the score estimate \(\nabla_z \log p(z_t|c)\) used in the reverse diffusion process. 

Early training-based approaches learn explicit conditional modules, such as classifiers or condition-dependent score estimators, to provide guidance~\cite{sddiffusion,ppt,brushnet}. Although these methods achieve strong controllability, they require additional training for each new condition, which is computationally costly and limits adaptability across tasks.

More recently, a growing body of work has explored training-free guidance, in which the diffusion trajectory is modified directly without retraining any auxiliary networks. Representative approaches include DPS~\cite{dps} with posterior guidance, MPGD~\cite{mpdg} with manifold-aware sampling, and ADMM-Diff~\cite{admmdiff} with an ADMM-based conditional diffusion strategy. However, these methods inject external guidance signals directly into the score update during diffusion sampling, implicitly assuming that the guidance direction itself is compatible with the geometry learned by the pre-trained diffusion model. This assumption is often violated in adversarial settings, where task-driven gradients may push the diffusion trajectory toward unlikely or out-of-distribution regions, resulting in unstable sampling or visually implausible artifacts. In contrast, we propose Jacobian Vector Product Guidance (JVPG), which explicitly models the interaction between external adversarial gradients and the local geometry of the diffusion model. By modulating the guidance direction through the Jacobian vector product of the pre-trained diffusion model, JVPG reshapes adversarial updates to remain aligned with the learned diffusion geometry. This enables effective adversarial object generation that induces substantial depth distortion while preserving visual plausibility.

\section{Preliminaries}

\subsection{Score-Based Diffusion Models}
Let $z_0$ denote a clean data sample in latent space. The forward noising process gradually perturbs $z_0$ into a sequence of increasingly noisy variables $\{z_t\}_{t=1}^T$, while a score network $s_\theta(z_t,t)$ is trained to approximate the score function $\nabla_{z_t}\log p(z_t)$, that is, $s_\theta(z_t,t)\approx\nabla_{z_t}\log p(z_t)$~\cite{song1,song2}. The reverse dynamics of DDIM~\cite{ddim} are given by
\begin{equation}
\begin{aligned}
z_{t-1}
=&\;
\frac{1}{\sqrt{{\alpha}_t}}\; z_t+
\sigma_t \, \epsilon 
+
\Bigg(
\frac{\left(1-\bar{\alpha}_t\right)}{\sqrt{{\alpha}_t}}
-
\sqrt{\left(1-\bar{\alpha}_{t-1}-\sigma_t^2\right)}
\\
&
\cdot \sqrt{
\left(1-\bar{\alpha}_t\right)}
\Bigg)
\nabla_{z_t}\log p(z_t),
\end{aligned}
\label{ddim}
\end{equation}
where $\epsilon \sim \mathcal{N}(0,I)$ is standard Gaussian noise, $\alpha_t \in (0,1)$ is a prescribed noise schedule with
$\bar{\alpha}_t := \prod_{i=1}^t \alpha_i$,
and $\sigma_t \ge 0$ is the sampling noise scale.

\subsection{Conditional Diffusion via Score Modification}

To enable diffusion models to accommodate diverse downstream objectives, it is necessary to introduce conditional mechanisms that allow controllable generation. Conditional diffusion models achieve this goal by incorporating external conditions into the sampling dynamics. As shown in prior work~\cite{cg}, conditioning on a variable $c$ can be formulated by modifying the score function to estimate the gradient of the conditional density \( \nabla_{z_t} \log p(z_t \mid c) \).

By applying Bayes' rule,
$
p(z_t \mid c) = \frac{p(z_t)\, p(c \mid z_t)}{p(c)},
$
the conditional score can be decomposed into two additive components:
\begin{equation}
\nabla_{z_t} \log p(z_t \mid c)
= \nabla_{z_t} \log p(z_t)
+ \nabla_{z_t} \log p(c \mid z_t).
\label{eq:conditional_score}
\end{equation}

The first term can be directly obtained from a pre-trained score network $s_\theta(z_t,t)$. In contrast, the second term encodes the influence of the condition and plays a central role in enabling conditional generation. This term can be interpreted as a guidance signal that steers the sampling trajectory toward regions of the latent data space consistent with the imposed condition $c$. Existing classifier-based guidance approaches~\cite{cg,energy2} approximate this conditional gradient by training a time-dependent classifier to estimate $\nabla_{z_t} \log p(c \mid z_t)$, which is then injected into the diffusion dynamics to bias the generation process.

\subsection{Energy-Based Guidance}

However, training an additional conditional network for adversarial guidance is often impractical in our setting, as it requires task-specific supervision and substantially increases training cost, while also limiting the generality of the framework across different attack objectives and target models. To avoid these issues, we adopt a training-free formulation in which $p(c \mid z_t)$ is defined implicitly through an energy function~\cite{energy1,energy2,freedom}:
\begin{equation}
p(c\mid z_t)=
\frac{exp\{-\gamma \, g_\theta(c,z_t)\}}{Z},
\label{eq:energy}
\end{equation}
where $\gamma$ controls the guidance strength and $Z>0$ denotes a normalizing constant. $g_\theta(c, z_t)$ quantifies the compatibility between the noisy latent variable $z_t$ and the condition $c$. Under this formulation, lower energy values correspond to higher consistency with the imposed condition, while configurations that violate the condition incur larger energy penalties.

Compared with time-dependent conditional networks that directly operate on noisy latents, many condition-related similarity or distance functions are naturally defined on clean data representations. Such functions $h_\theta$ provide time-independent measures of compatibility between a condition $c$ and a clean latent variable $z_0$, but cannot be directly evaluated on the noisy latent $z_t$ encountered during diffusion sampling. To address this mismatch, we follow prior work~\cite{freedom} and approximate the clean latent using the posterior mean conditioned on $z_t$~\cite{tw}:

\begin{equation}
z_{0|t}
=
\frac{1}{\sqrt{\bar{\alpha}_t}}
\Big(z_t + (1-\bar{\alpha}_t)\,s_\theta(z_t,t)\Big),
\label{eq:posterior}
\end{equation}
which allows the energy to be approximated by
\begin{equation}
\nabla_{z_t} \log p(c \mid z_t)
\propto
-\nabla_{z_t}g_\theta(c,z_t)
\approx
-\nabla_{z_t}h_\theta(c,z_{0|t}).
\label{eq:compat}
\end{equation}

Combining Eq.\eqref{ddim}, Eq.\eqref{eq:conditional_score} and Eq.\eqref{eq:compat}, the conditional sampling can be written as:
\begin{equation}
\begin{aligned}
z_{t-1}
=&\;
\frac{1}{\sqrt{{\alpha}_t}}\; z_t+
\sigma_t \, \epsilon 
+
\Bigg(
\frac{\left(1-\bar{\alpha}_t\right)}{\sqrt{{\alpha}_t}}
-
\sqrt{\left(1-\bar{\alpha}_{t-1}-\sigma_t^2\right)}
\\
&
\cdot \sqrt{
\left(1-\bar{\alpha}_t\right)}
\Bigg)\Bigg(s_\theta(z_t,t)-\gamma \, \nabla_{z_t}h_\theta(c,z_{0|t})\Bigg).
\end{aligned}
\label{condition_sample}
\end{equation}

\begin{figure*}[!t]
\centering
\includegraphics[width=\textwidth]{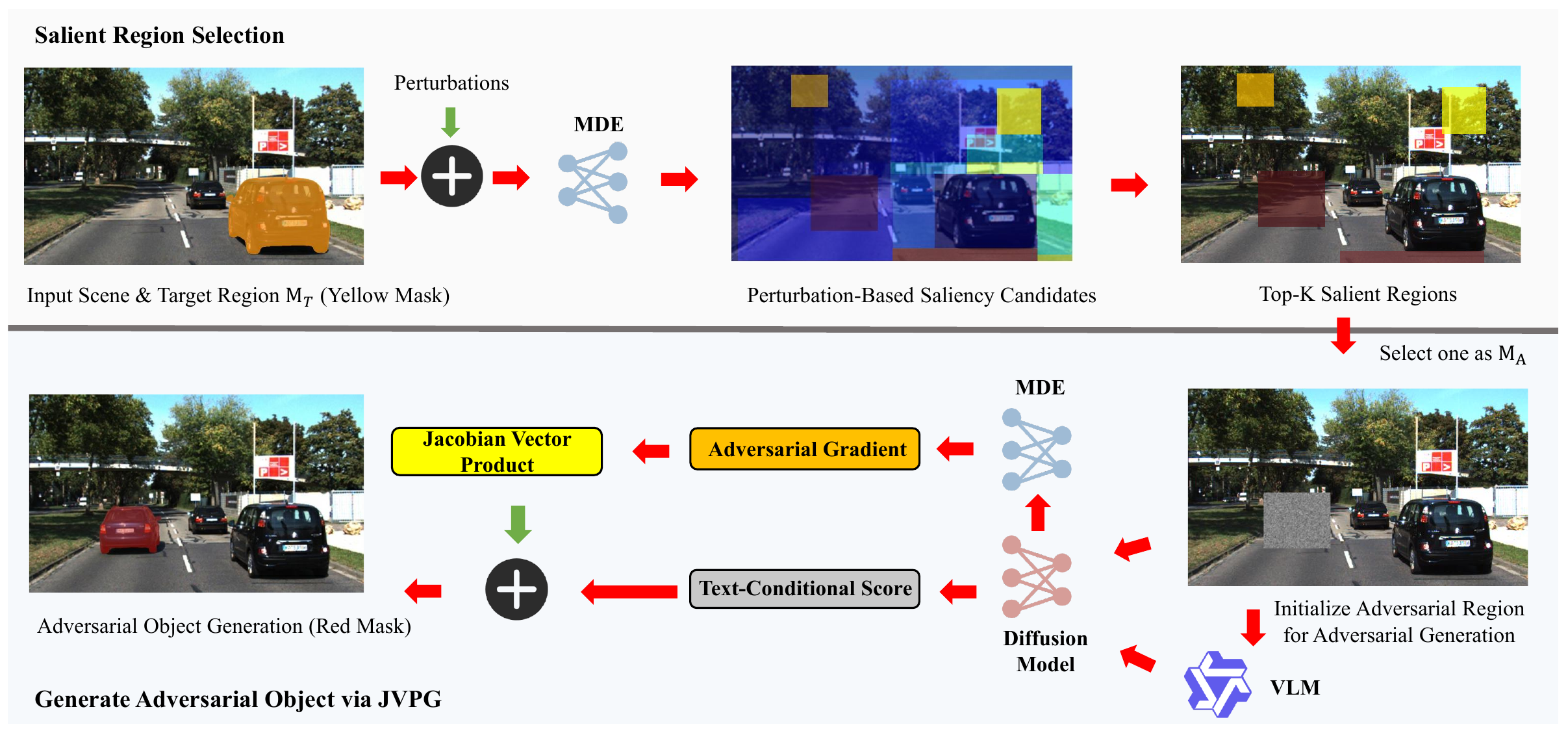}
\caption{Overview of the proposed generative adversarial attack framework. The pipeline first performs Salient Region Selection by injecting perturbations into image regions, ranking their influence on the MDE model, and selecting the top-$K$ most vulnerable regions. In the second stage, a diffusion-based generator produces a scene-consistent adversarial object at the selected region, where Jacobian Vector Product Guidance (JVPG) injects adversarial gradients into the diffusion trajectory while preserving text-conditional semantics and visual realism, ultimately inducing substantial depth shifts in the MDE output.}
\label{overview}
\end{figure*}

\section{METHODOLOGY}
\label{sec:methodology}

\subsection{Problem Formulation}
\label{sec:problem_formulation}

In autonomous driving scenarios, an on-board system typically deploys a MDE model $f$ that predicts a depth map from a monocular RGB image $x\in\mathbb{R}^{3\times h\times w}$. We define two binary masks over the image:

\begin{itemize}
    \item Target region mask $M_T\in\{0,1\}^{h\times w}$, indicating the region whose depth estimation the attack aims to alter.
    \item Adversarial object mask $M_A\in\{0,1\}^{h\times w}$, indicating the region where adversarial content can be inserted.
\end{itemize}

Masking is performed via the Hadamard product $\odot$. Given an adversarial object $\mathcal{A}\in\mathbb{R}^{3\times h\times w}$, the adversarial scene is constructed as
$
z = x\odot(1 - M_A)+ \mathcal{A}  \odot  M_A,
$
so that $\mathcal{A} = z  \odot  M_A$ represents the inserted object. For convenience, the notation $f(x)\odot M_T$ is abbreviated as $f_{M_T}(x)$ and adversarial objective $L_{\text{adv}}$, i.e., depth difference, is measured through
\begin{equation}
L_{\text{adv}}(x,z,M_T)
=
\big\| f_{M_T}(z) - \lambda\ \cdot f_{M_T}(x) \big\|_2^2,
\label{eq:La_final}
\end{equation}
where $\lambda$ is a scaling factor that regulates the desired magnitude of depth deviation.

Existing physical attacks~\cite{phmde,advrm} optimize the appearance of a fixed-style patch within a predetermined $M_A$, forcing the adversary to rely on handcrafted textures and constrained spatial locations. This severely limits realism, semantic compatibility, and the ability to deploy attacks at arbitrary positions.

Instead, we adopt a two-stage attack framework for generating contextually plausible adversarial objects at arbitrary locations in the scene. The overall system, shown in Fig.~\ref{overview}, consists of:
(1) \emph{Salient Region Selection}, and
(2) \emph{Adversarial Object Generation via JVPG}.

In Stage~(1), the scene is partitioned into candidate patches, each representing a potential insertion region. For every patch, we estimate its adversarial saliency by measuring the depth perturbation induced by localized disturbances.
These saliency scores are then ranked to obtain the top-$k$ insertion regions $M_A$ that are most influential to the target depth prediction.

Given the selected regions, Stage~(2) performs conditional adversarial generation. We introduce a Jacobian Vector Product Guidance (JVPG) mechanism that first computes adversarial gradients to induce depth misestimation on the target region, and then refines these gradients by adjusting their update directions via the Jacobian vector product with the pre-trained score network before injecting them into the diffusion trajectory to generate adversarial content.
This enables the diffusion model to generate object appearances that are both adversarially effective and visually coherent with the surrounding scene.

This formulation offers two key advantages over patch-based methods. First, it places no restriction on the spatial support of the adversarial region: the model can generate contextually coherent adversarial objects for any $M_A$ provided at Salient Region Selection. 
Second, the object’s final appearance emerges from the interplay between diffusion prior and adversarial objective $L_{\text{adv}}$, rather than being hand-designed, yielding significantly greater realism and more effective adversarial depth shifts.

\subsection{Salient Region Selection}   

\begin{figure}[!t]
\centering
\includegraphics[width=\linewidth]{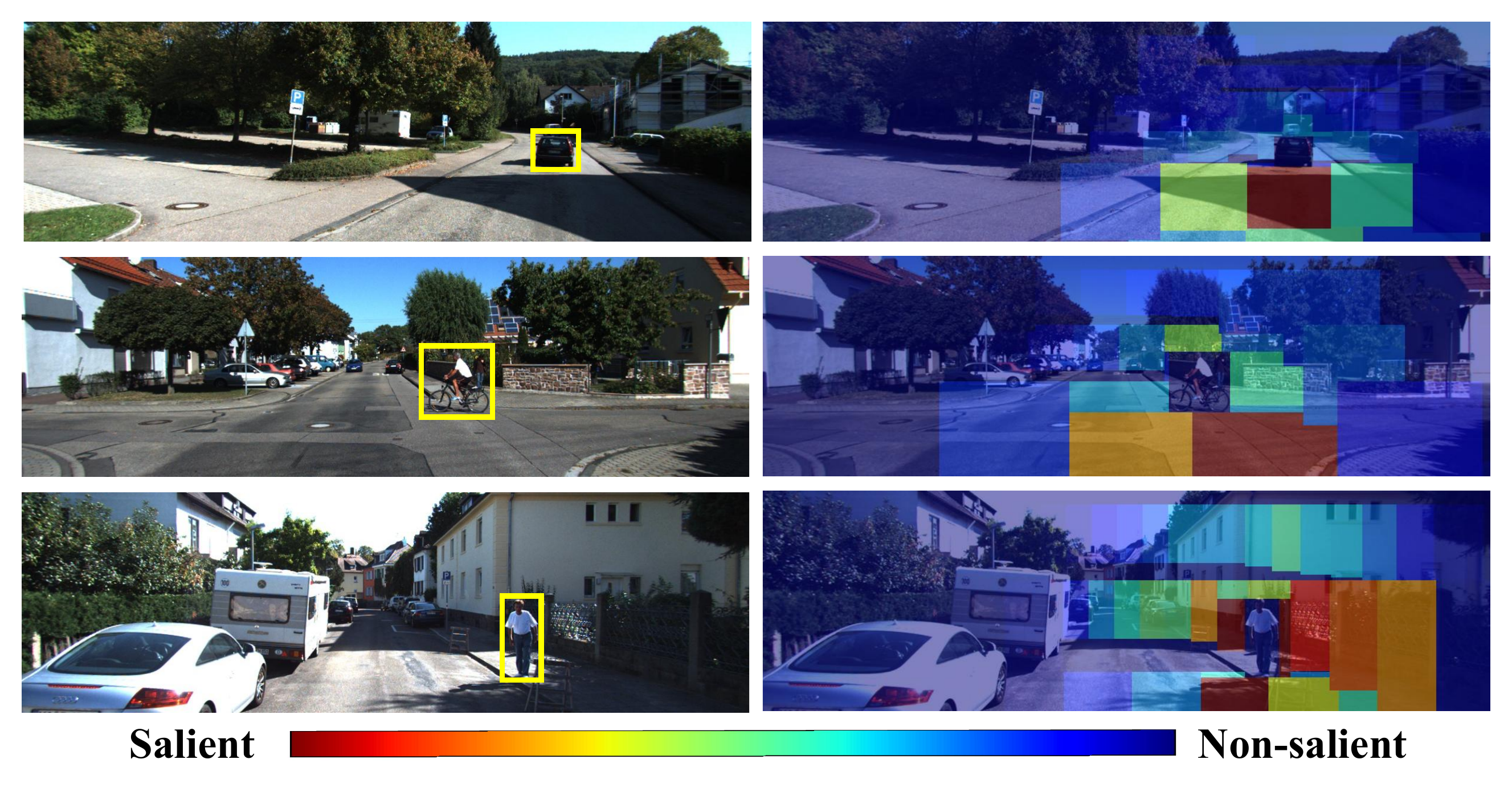}
\caption{Visualization of salient region estimation across diverse driving scenes. 
The yellow bounding box denotes the target object mask $M_T$. 
Warmer colors indicate regions with higher saliency scores, which exert stronger influence on the depth predicted within $M_T$ and are therefore prioritized for adversarial object insertion, whereas cooler colors correspond to non-salient regions.}
\label{salient}
\end{figure}

Inspired by prior robustness analyses of DNNs~\cite{sradv,advrm}, we observe that different spatial regions in an image contribute unequally to the depth prediction of a target object. To characterize this non-uniform influence, we introduce a \emph{Salient Region Selection} module that identifies the regions most critical to the MDE output.

Given an input image $x$, we first partition it into a set of $N$ candidate patches whose spatial sizes are adaptively determined from the depth and geometric extent of the target mask $M_T$. The goal is to quantify how sensitive the target prediction $f_{M_T}(x)$ is to perturbations restricted within each patch. This ranking allows adversarial generation to focus on regions to which the MDE model is inherently most vulnerable.

Let $M_P$ denote the binary mask of a candidate patch. We evaluate the local perturbation impact of $f_{M_T}$ using the objective as
\begin{equation}
\mathcal{L}(u)
=
\big\|f_{M_T}(x+u)-f_{M_T}(x)\big\|_2,
\label{eq:srs_loss}
\end{equation}
where $u$ is a perturbation supported only within $M_P$. We update $u$ by gradient ascent to increase the target loss $\mathcal{L}(u)$, yielding
\begin{equation}
u \leftarrow u + \eta \frac{\nabla_u \mathcal{L}(u)}{\|\nabla_u \mathcal{L}(u)\|_2},
\end{equation}
where $\eta$ is the step size. This update corresponds to a first-order gradient ascent step within the patch-constrained subregion.

After optimization, the saliency of region $i$ is defined as
\begin{equation}
\phi(i)
=
f_{M_T}(x+u_i) - f_{M_T}(x),
\label{eq:srs_phi}
\end{equation}
where $u_i$ is the optimized perturbation within patch $i$. A larger $\phi(i)$ indicates that modifying this region leads to a stronger shift in depth prediction, revealing it as a more influential and potentially vulnerable location. Algorithm~\ref{alg:salient_region} summarizes the procedure, and Fig.~\ref{salient} provides qualitative visualizations of the resulting saliency maps.

Overall, this module leverages first-order sensitivity analysis to identify regions where perturbations produce the largest changes in the target depth prediction. These salient regions serve as high-value candidates for the 
subsequent adversarial generation stage.

\begin{algorithm}[t]
\caption{Salient Region Selection}
\label{alg:salient_region}
\begin{algorithmic}[1]
\Require Image $x$, MDE model $f$, target mask $M_T$, iterations $T$, step size $\eta$, top-$k$
\Ensure Ranked salient regions
\State Generate patch masks $\mathcal{C}$ adaptively according to the depth and spatial extent of $M_T$
\State Remove or trim patches that overlap with $M_T$
\State Compute baseline depth $D = f_{M_T}(x)$
\For{each patch mask $M_P \in \mathcal{C}$}
    \State Initialize perturbation $u \gets 0$
    \For{$t = 1$ to $T$}
        \State $g \gets \nabla_u \| f_{M_T}(x+u) - D \|_2$
        \State $u \leftarrow u + \eta \, \frac{g}{\|g\|_2}$
    \EndFor
    \State $\mathrm{Score}(M_P) \gets f_{M_T}(x+u) - D$
\EndFor
\State Rank all patches by $\mathrm{Score}(M_P)$ in descending order
\State \Return top-$k$ salient regions
\end{algorithmic}
\end{algorithm}

\subsection{Generate Adversarial Object via JVPG}

\begin{figure*}[!t]
\centering
\includegraphics[width=\textwidth]{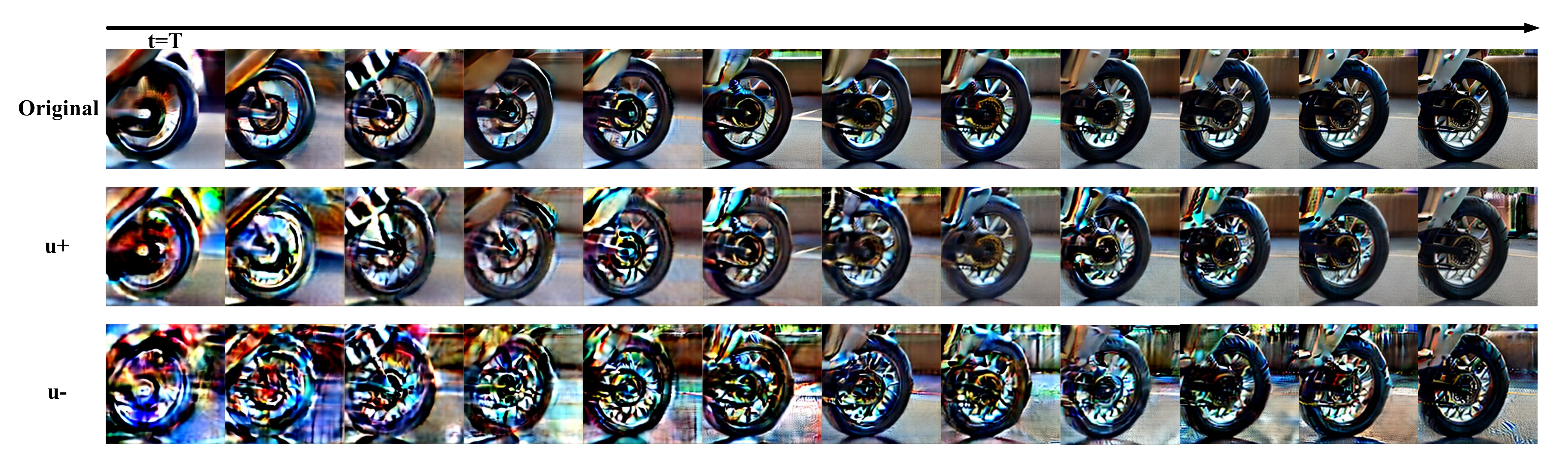}
\caption{Comparison of denoising trajectories under different Jacobian singular directions. The first row shows the original diffusion trajectory. The second row applies perturbations along the dominant singular direction $u^+$, which preserves coherent semantic structures. The third row applies perturbations along the smallest singular direction $u^-$, resulting in disordered, non-semantic artifacts. These visualizations highlight that $u^+$ corresponds to meaningful generative directions, whereas $u^-$ drives the diffusion process away from semantic consistency.}
\label{jvpgvis}
\end{figure*}

Given a target region $M_T$ and an adversarial insertion region $M_A$, our goal is to generate a background-consistent adversarial object $\mathcal{A}$ that perturbs the depth prediction $f_{M_T}(x)$. To this end, we leverage the generative prior of a pre-trained text-guided diffusion model and generate a full adversarial scene $z$, from which the inserted object is obtained as $\mathcal{A} = z \odot M_A$. Generating the full scene $z$ further allows the diffusion model to exploit the surrounding image structure when synthesizing the adversarial object, enabling $\mathcal{A}$ to adapt naturally to its local context and remain visually plausible. 

Under this formulation, the generation process is weakly conditioned by a coarse textual description $c_{\text{text}}$, obtained from a pre-trained Vision--Language Model (VLM)~\cite{qwen}, which specifies only the semantic category of the object. Beyond this high-level semantic constraint, the object’s final appearance is not determined by scene context alone, but emerges from the interaction between the diffusion prior and the adversarial objective $L_{\text{adv}}$, which jointly shape the geometry, shading, and texture required to induce depth misestimation in $M_T$.

In practice, the text condition $c_{\text{text}}$ is obtained from a pre-trained VLM~\cite{qwen}, which takes the original image $x$ and the insertion region $M_A$ as input and returns a concise description of a plausible object for that location. This step is necessary because, in the absence of an explicit semantic cue, diffusion models often revert to generating 
dominant visual modes seen during training~\cite{brush2prompt}, leading to objects that are visually salient and thus insufficiently stealthy for physical adversarial attacks. By supplying a weak semantic prior, $c_{\text{text}}$ constrains the generative process toward realistic and context-appropriate objects while leaving the fine-grained adversarial appearance to be determined by $L_{\text{adv}}$.

On the other hand, diffusion models natively support a set of pre-trained conditioning signals~\cite{sddiffusion,controlnet,ppt,brushnet}, denoted collectively as $c_{\text{i}}$. Specifically, $c_i = \{x, M_A, c_{\text{text}}\}$, such as textual prompts $c_{\text{text}}$ are encoded by the model’s text encoder and injected via cross-attention layers. Importantly, these intrinsic conditions are already learned during diffusion training and require no additional modeling.

In contrast, the adversarial objective $L_{\text{adv}}$, which specifies how the generation should perturb the MDE output, is an extrinsic constraint that the 
diffusion model has never encountered during training. As derived in Eq.~\eqref{condition_sample}, its effect is incorporated 
through the energy-gradient term $\nabla_{z_t} h_\theta(c, z_{0|t}):= \nabla_{z_t} L_{\text{adv}}(x, z_{0|t}, M_T)$.

In practice, we observe that directly injecting adversarial gradients during diffusion sampling can drive the sampling trajectory toward out-of-distribution directions that are poorly supported by the pre-trained score network. Such a mismatch often manifests as visually implausible textures and noticeable artifacts in the generated objects, indicating that naive gradient injection fails to respect the geometric structure learned by the diffusion model.

To analyze how external perturbations interact with the intrinsic geometry encoded by the score network, we adopt a Jacobian-based perspective. Prior analysis in~\cite{jvp1} shows that the Jacobian of the score function captures the local curvature of the underlying data density. To further examine the properties of the Jacobian, we perform a singular value decomposition of the Jacobian, $J = U \Sigma V^\top$, and identify the left singular vector $u^+$ corresponding to the largest singular value, as well as $u^-$ corresponding to the smallest one. As illustrated in Fig.~\ref{jvpgvis}, perturbations injected along $u^+$ preserve coherent semantic content, whereas perturbations along $u^-$ tend to introduce disordered and non-semantic artifacts.

Motivated by these observations, we propose \emph{Jacobian Vector Product Guidance (JVPG)}, which refines the adversarial gradient by modulating its projection onto Jacobian directions. Specifically, JVPG amplifies the external perturbation along the semantic direction like $u^+$ while suppressing the component aligned with the non-semantic direction like $u^-$, thereby steering the diffusion updates toward perceptually plausible adversarial objects rather than noisy artifacts.
At each timestep, we compute the adversarial perturbation on the noisy state
\begin{equation}
\delta 
=
\nabla_{z_t} L_{\text{adv}}(x,z_{0|t},M_T).
\label{eq:jvpg_delta}
\end{equation}

In practice, explicitly computing the full Jacobian of the score network is computationally expensive. To enable an efficient approximation, we locally linearize the score function around $z_t$, yielding
\begin{equation}
\begin{aligned}
s_\theta(z_t - \delta, t \mid c_i)
&\approx
s_\theta(z_t, t \mid c_i)
-
J_{s_\theta}(z_t, t \mid c_i)\,\delta,
\end{aligned}
\label{eq:jvpg_linear}
\end{equation}
where $J_{s_\theta}(z_t,t \mid c_i)$ denotes the Jacobian of the score network with respect to $z_t$. The Jacobian vector product $J_{s_\theta}\delta$ describes how the adversarial direction is transformed by the local score geometry. In particular, rather than uniformly amplifying all perturbation components, the Jacobian selectively emphasizes directions aligned with the semantic subspace captured by the score, while suppressing those lying in non-semantic or noisy directions.

Substituting Eq.~\eqref{eq:jvpg_linear} into the conditional update Eq.~\eqref{condition_sample} yields our JVPG-guided reverse step:
\begin{equation}
\begin{aligned}
z_{t-1}
=&\;
\frac{1}{\sqrt{{\alpha}_t}}\; z_t+
\sigma_t \, \epsilon 
+
\Bigg(
\frac{\left(1-\bar{\alpha}_t\right)}{\sqrt{{\alpha}_t}}
-
\sqrt{\left(1-\bar{\alpha}_{t-1}-\sigma_t^2\right)}
\\
&
\cdot \sqrt{
\left(1-\bar{\alpha}_t\right)}
\Bigg)\Bigg(s_\theta(z_t,t\mid c_i)-\gamma \, J_{s_\theta}(z_t,t \mid c_i)\,\delta)\Bigg).
\end{aligned}
\label{eq:jvpg_update}
\end{equation}

\begin{algorithm}[t]
\caption{Jacobian Vector Product Guidance (JVPG)}
\label{alg:jvpg}
\begin{algorithmic}[1]
\Require Target mask $M_T$, adversarial region $M_A$, image $x$, MDE model $f$, score network $s_\theta$, textual description $c_\text{text}$, diffusion steps $T$, noise schedule $\alpha_t$ 
\Ensure Adversarial scene $z$ and object $\mathcal{A}$
\State Sample $z_T \sim \mathcal{N}(0,I)$ \Comment{Initialize noisy state}
\For{$t = T$ down to $1$}
    \State $z_{0|t} \gets \frac{1}{\sqrt{\bar{\alpha}_t}}\big(z_t + (1-\bar{\alpha}_t)s_\theta(z_t,t \mid c_i)\big)$
    \State $\delta \gets \nabla_{z_t} L_{\text{adv}}(x,z_{0|t},M_T)$ 
    \State $\mathbf{z}_t^{\mathrm{adv}} \gets \mathbf{z}_t + \delta$ 
    \State Update Jacobian vector product via Eq.~\eqref{eq:jvpg_linear}
    \State Update sampling via Eq.~\eqref{eq:jvpg_update}
\EndFor
\State $z = z_0$ \Comment{Final adversarial scene}
\State Extract object: $\mathcal{A} = z \odot M_A$
\State \Return $z, \mathcal{A}$
\end{algorithmic}
\end{algorithm}

Algorithm~\ref{alg:jvpg} summarizes the procedure of JVPG. By leveraging the Jacobian vector product, JVPG provides an implicit, timestep-adaptive modulation of the adversarial influence that respects the geometry of the pre-trained diffusion model. This ensures that the adversarial object generates along semantically consistent directions of the generative manifold while still exerting influence on the depth predictions. Importantly, since the guidance operates directly on the generative process, the method naturally adapts to any insertion region $M_A$ at inference time, enabling stealthy adversarial objects.

\section{EXPERIMENTS}
\subsection{Experimental Setup}

\subsubsection{Dataset}
Existing physical attacks on MDE, such as~\cite{phmde}, have commonly adopted subsets of the KITTI dataset~\cite{kitti}. However, many selected targets, such as buildings or trees, do not directly influence driving decisions, and perturbing their depth provides limited insight into the practical safety risks faced by autonomous vehicles. Other works, such as AdvRM~\cite{advrm}, evaluate attacks in highly idealized straight-road settings using synthetically inserted targets and lack publicly released experimental details.

To provide a fair, realistic, and safety-oriented evaluation, we construct a unified benchmark derived from real KITTI driving sequences. Using an optical-flow-based selection strategy, we extract 459 diverse scenes covering both straight-road and roadside scenarios. In contrast to prior synthetic settings~\cite{advrm}, our benchmark preserves the original scene geometry and imaging conditions. Notably, our benchmark is approximately $4.5\times$ larger than the existing benchmark~\cite{advrm} and contains over $3\times$ more categories of common traffic objects, covering a broad spectrum of vehicles, roadside infrastructure, and traffic-related entities commonly encountered in real-world driving environments. This combination of increased data scale and object diversity enables a more comprehensive and realistic assessment of adversarial robustness. The benchmark will be publicly released to facilitate future research.

Specifically, we employ Grounded SAM~\cite{ren2024grounded} to annotate common traffic objects, including both on-road actors (e.g., cyclists and moving vehicles) and off-road entities that constitute hidden hazards, such as parked cars and roadside pedestrians. Although these objects may not lie within the drivable area at the moment a frame is captured, their depth estimation
is crucial: underestimated distance to a parked vehicle may delay braking if it suddenly re-enters the lane, and inaccurate depth for a roadside pedestrian may hinder timely collision-avoidance planning. By focusing on such safety-relevant targets, our benchmark better reflects the depth
estimation challenges that autonomous vehicles encounter in real-world operation.

\subsubsection{Implementation Details}

We employ the pre-trained PowerPaint-v2 model~\cite{ppt,brushnet} as the diffusion backbone for adversarial object generation, and use Qwen3-VL~\cite{qwen} as the VLM to provide the text condition $c_{\text{text}}$. Unless otherwise specified, the number of selected regions is set to $k=4$, allowing up to four adversarial objects to be inserted in a single scene.

For adversarial guidance, we set $\lambda = 2$ in Eq.\eqref{eq:La_final} for all experiments, which empirically yields a stable trade-off between attack strength and visual plausibility.

\subsubsection{Evaluation Metrics}
To evaluate the effectiveness of the proposed method, we employ two complementary metrics. In particular, we adopt the CLIP-Score (C-S)~\cite{clip} to assess the perceptual realism and semantic correctness of the generated adversarial objects. While Fréchet Inception Distance (FID)~\cite{fid} is a standard metric for generative models and measures the distributional distance between generated images and real images from a target class, it requires access to a representative ground-truth image distribution. In our setting, however, the objective is not to match a predefined dataset distribution, but to verify whether the generated objects are visually realistic and semantically consistent with the intended object category. C-S provides a reference-free alternative by directly measuring the semantic alignment between the generated adversarial objects and their corresponding textual descriptions, making it better suited for evaluating object-level realism in our open-vocabulary generation setting.

In addition, we use the Mean Relative Shift Ratio (MRSR)~\cite{advrm}, denoted as $\xi_r$, to quantify the depth shift of the target object after the attack.
Specifically, $\xi_r$ is defined as
\begin{equation}
\xi_r \;=\; 
\frac{\sum \big( f_{M_T}(z) - f_{M_T}(x) \big)}
     {\sum f_{M_T}(x)},
\label{eq:mrsr}
\end{equation}
where the summation is taken over all pixels within $M_T$. A larger $\xi_r$ indicates a stronger adversarial effect, corresponding to a more pronounced deviation in the estimated depth of the target object. To remain consistent with common camera acquisition pipelines, all evaluations are conducted on JPEG-encoded images.

\subsection{Dataset Simulation}

\begin{table*}[!t]
\centering
\caption{Comparison of attack effectiveness between AdvRM and our method across multiple MDE models. The ``Regions'' column denotes the number of adversarial insertion regions. Performance is measured using MRSR.}
\label{tab:method_region}

\begin{tabularx}{\textwidth}{*{7}{>{\centering\arraybackslash}X}}
\toprule 
\textbf{Method} & \textbf{Regions} & \textbf{MonoDepth2}~\cite{monodepth2} & \textbf{DepthHints}~\cite{depthhints} & \textbf{ManyDepth}~\cite{manydepth} &
\textbf{MonoDEVSNet}~\cite{MonoDEVSNet}& \textbf{DepthAnything}~\cite{depthanything} \\
\midrule 
AdvRM~\cite{advrm} & / & 0.31 & 0.21 & 0.12 &0.22& 0.04 \\

\multirow{5}{*}{\textbf{Ours}} & 1 & 0.17 & 0.14 & 0.14&0.14 & 0.12 \\
                     & 2 & 0.32 & 0.26 & 0.25&0.28 &  0.18\\
                     & 3 & 0.46 & 0.36 & 0.35&0.41 & 0.24 \\
                     & 4 & \textbf{0.59} & \textbf{0.46} & \textbf{0.43}&\textbf{0.52} & \textbf{0.29} \\
\bottomrule 
\end{tabularx}
\end{table*}

\begin{figure*}[!t]
\centering
\includegraphics[width=\textwidth]{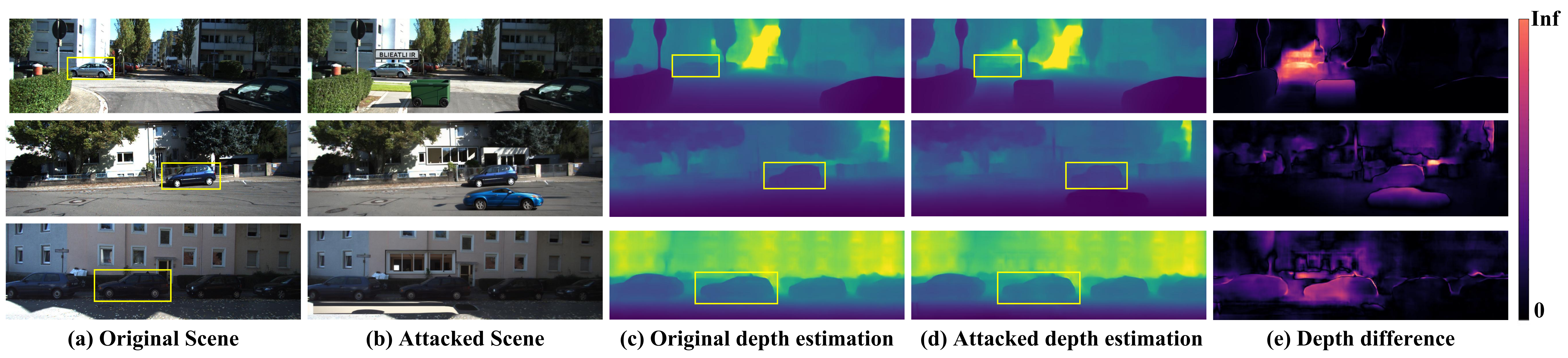}
\caption{Qualitative visualization of our generative adversarial object attack and its impact on MDE. From left to right: (a) original RGB scene, (b) adversarial scene with the generated object inserted, (c) predicted depth map for the original scene, (d) predicted depth map for the adversarial scene, and (e) depth difference map. The yellow bounding box marks the target region $M_T$, where the induced depth shift is evaluated. Brighter regions in the fifth column indicate larger depth deviations, highlighting that our method induces significant depth shifts while preserving realistic appearance in the digital domain.}
\label{visdiff}
\end{figure*}

We evaluate the proposed attack on our benchmark using several mainstream MDE models trained on the KITTI dataset. Specifically, we consider four CNN-based models, MonoDepth2~\cite{monodepth2}, DepthHints~\cite{depthhints}, ManyDepth~\cite{manydepth} and MonoDEVSNet~\cite{MonoDEVSNet}, as well as the transformer-based DepthAnything~\cite{depthanything}. These models differ substantially in network architecture, training objectives, and data utilization strategies, thereby providing a diverse set of victim models for evaluating the generality of the proposed attack.

To ensure a fair and conceptually consistent comparison, we distinguish between patch-based and generative attack paradigms. AdvRM~\cite{advrm} is a state-of-the-art patch-based adversarial method that optimizes a fixed-location perturbation, whereas our approach performs adversarial object generation and naturally supports inserting multiple objects at different spatial locations. Accordingly, we first apply the proposed Salient Region Selection algorithm to identify influential regions in each scene. Adversarial objects are then generated and inserted into the top-$k$ selected regions. Under this setting, we compare the attack effectiveness of our method with AdvRM.

As shown in Table~\ref{tab:method_region}, our method consistently outperforms AdvRM across all evaluated MDE models. More importantly, the attack effectiveness increases monotonically as the number of insertion regions grows, revealing a fundamental advantage of generative adversarial attacks over patch-based methods.  This multi-region capability allows the attack to influence the global depth geometry inferred by the MDE model, leading to substantially stronger and more stable depth misestimation. When four adversarial regions are used, our method achieves an average MRSR of 0.46 across five models, significantly exceeding AdvRM, which is limited to a single fixed patch and attains only 0.18 on average. These results demonstrate that generative attacks with multiple regions provide a fundamentally different attack capability: instead of perturbing a single local area, they enable coordinated manipulations over multiple influential regions, resulting in stronger and more stable depth misestimation than traditional patch-based approaches. As shown in Fig.~\ref{visdiff}, our generated adversarial objects remain photorealistic and semantically coherent with the surrounding scene. At the same time, they induce pronounced depth shifts within the target region $M_T$, demonstrating that the proposed method can simultaneously achieve high attack potency and visual plausibility in the digital domain.

\begin{table*}[!t]
\centering
\caption{Performance comparison between our JVPG and several mainstream training-free guidance methods, including DPS, MPDG, and ADMM-Diff. All methods use the same number of adversarial regions. Performance is measured using MRSR and C-S.}
\label{tab:ours_vs_guidance_img}
\begin{tabularx}{\textwidth}{*{6}{>{\centering\arraybackslash}X}}
\toprule 
\textbf{Method}  & \textbf{MonoDepth2}~\cite{monodepth2} & \textbf{DepthHints}~\cite{depthhints} & \textbf{ManyDepth}~\cite{manydepth} & 
\textbf{MonoDEVSNet}~\cite{MonoDEVSNet}& \textbf{DepthAnything}~\cite{depthanything} \\
 &MRSR$\uparrow$/C-S$\uparrow$&MRSR$\uparrow$/C-S$\uparrow$&MRSR$\uparrow$/C-S$\uparrow$&MRSR$\uparrow$/C-S$\uparrow$&MRSR$\uparrow$/C-S$\uparrow$ \\
\midrule 
DPS~\cite{dps}  & 0.49/21.68 & 0.11/21.60 & 0.42/21.60&0.27/21.52 & 0.18/21.56 \\

MPDG~\cite{mpdg}  & 0.30/21.53 & 0.07/21.29 & 0.28/21.52&0.14/21.31 & 0.14/21.43 \\

ADMM-Diff~\cite{admmdiff}  & 0.48/21.69 & 0.38/21.47 & 0.42/21.64&\textbf{0.52}/21.00 & 0.21/21.49 \\

\textbf{Ours}  & \textbf{0.59}/\textbf{22.22} & \textbf{0.46}/\textbf{22.13} & \textbf{0.43}/\textbf{22.24}&\textbf{0.52}/\textbf{22.34} & \textbf{0.29}/\textbf{22.04} \\
\bottomrule 
\end{tabularx}
\end{table*}

\begin{figure*}[!t]
 \centering
\includegraphics[width=\textwidth]{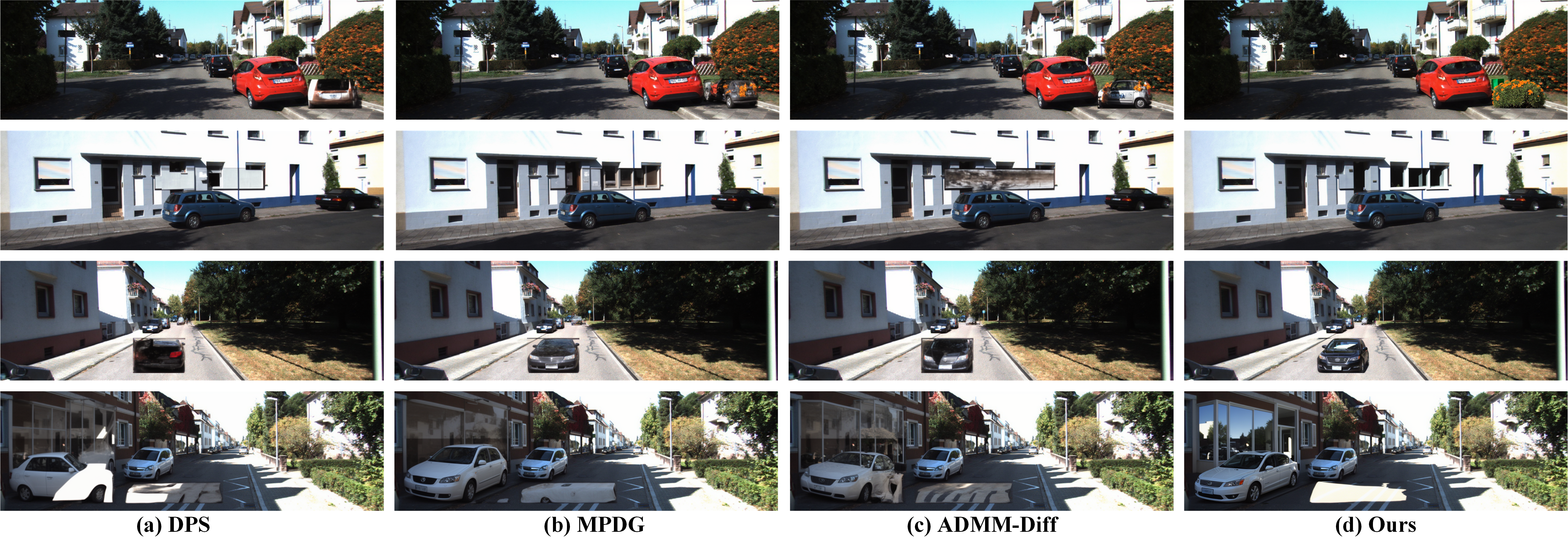}
\caption{Qualitative comparison of adversarial objects generated using different guidance strategies. Our JVPG guided generation produces more coherent textures and realistic geometry while maintaining strong attack potency.}
\label{generation}
\end{figure*}

This level of depth distortion is practically meaningful. For instance, a target object originally estimated at 20\,m may be perceived as 29.2\,m under attack. Such a depth overestimation can delay braking or alter distance-keeping behavior, potentially increasing collision risks in real driving scenarios.

To further validate the effectiveness of the proposed JVPG, we compare it with several representative training-free guidance strategies, including DPS~\cite{dps}, MPDG~\cite{mpdg}, and ADMM-Diff~\cite{admmdiff}. All methods operate on the same insertion regions $M_A$ and use the same text condition $c_{\text{text}}$ to ensure a controlled and fair comparison.

As shown in Table~\ref{tab:ours_vs_guidance_img} and Fig.~\ref{generation}, across all evaluated MDE models, JVPG consistently achieves the highest MRSR and C-S, outperforming the strongest competing method, ADMM-Diff, by a clear margin. Notably, JVPG improves attack strength without sacrificing semantic consistency, avoiding the typical trade-off observed in existing guidance strategies.

A more detailed analysis reveals distinct failure modes of existing guidance methods under the generative adversarial setting. DPS and ADMM-Diff often introduce noticeable visual artifacts, such as high-frequency noise patterns and geometrically inconsistent structures, as observed in Fig.~\ref{generation}. From an attack perspective, these artifacts partially explain the relatively high MRSR achieved by DPS and ADMM-Diff on certain models. The visually abrupt and unnatural patterns introduce strong and atypical depth cues, which can severely disrupt the depth estimation process and induce large depth shifts. However, such gains in attack effectiveness come at the expense of visual realism, indicating that high MRSR in these methods is often coupled with perceptually implausible perturbations.

In contrast, MPDG exhibits a different failure mode. As shown in Fig.~\ref{generation}, MPDG tends to generate objects that are only weakly integrated with the surrounding scene context. While the resulting images appear visually smooth and free of obvious artifacts, the inserted objects often lack semantic and geometric coherence with the environment, making them appear visually unnatural.
More importantly, such visually smooth but contextually disconnected objects exert only limited influence on scene-level depth reasoning. As a result, the generated content fails to significantly alter the depth structure perceived by the model, leading to lower MRSR values. This observation indicates that visual smoothness alone is insufficient for effective depth attacks. To meaningfully disrupt depth inference, adversarial objects must be coherently embedded into the scene geometry in a way that influences the model’s geometric reasoning.

Overall, these observations highlight a fundamental limitation of existing training-free guidance methods: they either prioritize attack strength at the cost of visual realism (DPS and ADMM-Diff), or preserve visual smoothness while lacking sufficient semantic and geometric influence on depth estimation (MPDG). JVPG resolves this dilemma by explicitly modulating adversarial gradients according to the diffusion geometry, enabling strong and stable depth distortion without introducing conspicuous artifacts.

\subsection{Ablation Experiments}
\subsubsection{Salient Region Selection}

\begin{table}[!t]
\centering
\caption{
Ablation study on region selection and adversarial guidance evaluated on MonoDepth2. We compare SRS and JVPG under different numbers of insertion regions. Performance is measured using the MRSR.
}
\label{tab:ab1}

\begin{tabularx}{\columnwidth}{
    >{\centering\arraybackslash}X
    *{4}{>{\centering\arraybackslash}X}
}
\toprule

\multirow{2}{*}{\centering \textbf{Method}} & \multicolumn{4}{c}{\textbf{Number of Regions}} \\
\cmidrule(lr){2-5}
 & 1 & 2 & 3 & 4 \\
\midrule
w/o SRS   & 0.04 & 0.07 & 0.10 & 0.12 \\
w/o JVPG  & 0.00 & -0.02 & -0.03 & -0.04 \\
\textbf{Ours}      & \textbf{0.17} & \textbf{0.32} & \textbf{0.46} & \textbf{0.59} \\
\bottomrule
\end{tabularx}

\end{table}

We first evaluate the effectiveness of the proposed Salient Region Selection (SRS) algorithm. For a fair comparison, we use MonoDepth2 as the victim MDE model and examine attack performance under different numbers of insertion regions. Two strategies are considered:
(i)~salient regions identified by SRS, and (ii)~random regions sampled uniformly across the image. Quantitative results in Table~\ref{tab:ab1} show that attacks conducted on SRS-selected regions consistently yield significantly larger depth shifts compared to random selection. This confirms that SRS successfully identifies the regions with the greatest influence on the target depth prediction, enabling more efficient adversarial object placement.

\subsubsection{Jacobian Vector Product Guidance}

We next assess the contribution of the proposed adversarial guidance mechanism. Specifically, we compare our JVPG with a baseline that performs standard inpainting without injecting adversarial gradients. As shown in Table~\ref{tab:ab1}, the baseline achieves only marginal or even negligible depth shifts, indicating that conventional inpainting alone is insufficient to mislead MDE models. In contrast, incorporating JVPG substantially amplifies the adversarial effect across all region counts, demonstrating that geometry-aware gradient modulation is crucial for generating objects that balance realism with strong attack potency.

\begin{figure*}[!t]
\centering
\includegraphics[width=\textwidth]{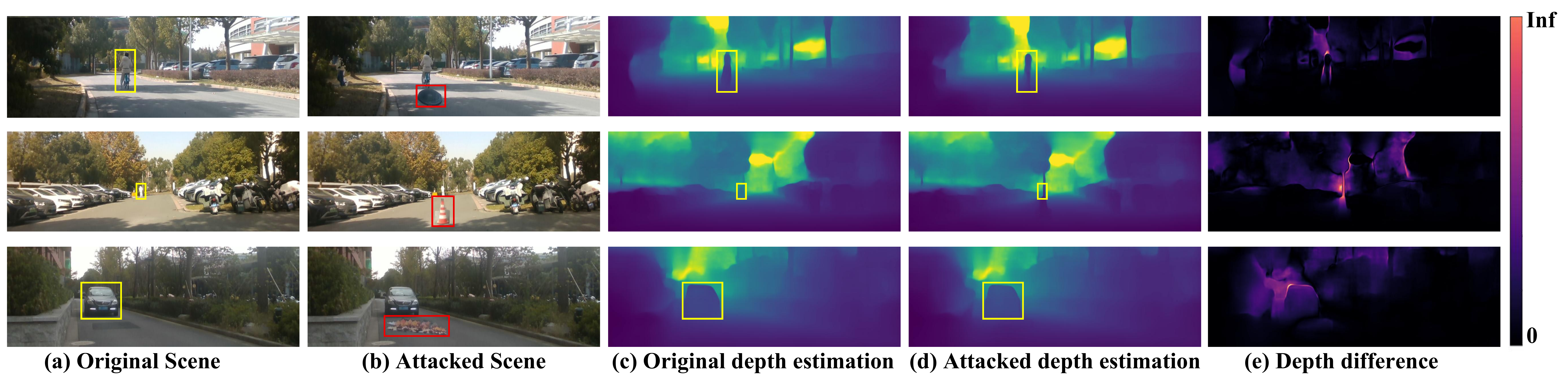}
\caption{Real-world deployment results for cyclist, pedestrian, and vehicle targets. 
The yellow box denotes the target region $M_T$, and the red box indicates the adversarial object region $M_A$. 
The adversarial objects are a printed metallic manhole cover (cyclist), a traffic cone (pedestrian), and a pile of fallen leaves (vehicle). 
Brighter regions in the depth-difference maps indicate larger depth deviations.}
\label{real}
\end{figure*}

\begin{figure}[!t]
\centering
\includegraphics[width=2.5in,,keepaspectratio]{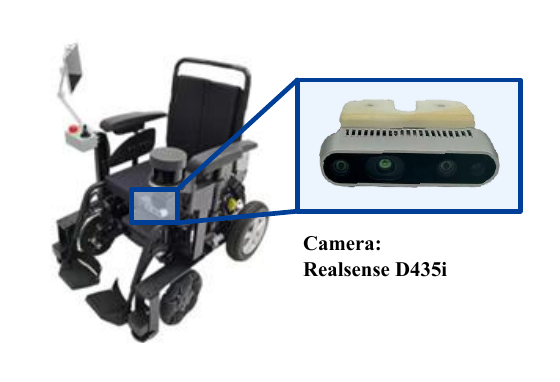}
\caption{Real-world experiment platform. We use a “JiaoLong” intelligent vehicle equipped with a RealSense D435i sensor as the autonomous driving platform to validate the real-world performance of our method.}
\label{platform}
\end{figure}

\begin{table}[!t]
\centering
\caption{MRSR of Digital and Physical Adversarial Attacks on Monodepth2 Evaluated on Real-World Driving Scenes.}
\label{tab:real}
\begin{tabularx}{\columnwidth}{*{3}{>{\centering\arraybackslash}X}}
\toprule 
\textbf{Scenario} & \textbf{Digital Domain} & \textbf{Physical Domain}\\
\midrule 
Cyclist & 0.21 & 0.18 \\
Pedestrian & 0.55 & 0.42 \\
Vehicle & 0.14 & 0.11 \\
\bottomrule 
\end{tabularx}
\end{table}

\subsection{Real-World Experiments}

We further validate the physical realizability of our attack using a “JiaoLong” intelligent vehicle~\cite{wheelchair,wang2025} as the autonomous driving platform, equipped with an Intel RealSense~D435i visual sensor, as shown in Fig.~\ref{platform}. All physical experiments use MonoDepth2 as the victim model to maintain consistency with our digital-domain evaluation.

Although our digital-domain analysis shows that multiple adversarial objects can be inserted simultaneously and exhibit stronger effects, in the physical world we evaluate only a single adversarial object due to the high cost of fabrication. In this experiment, we focus on whether the generated adversarial object can induce erroneous depth estimates under real sensor noise while maintaining visual stealthiness.

Following prior physical attacks~\cite{phmde,advrm}, we materialize the generated adversarial object by printing it and deploying it in real-world scenes. We consider three representative target categories commonly encountered in transportation scenarios: \textit{cyclist}, \textit{pedestrian}, and \textit{vehicle}. All experiments are carried out at a dedicated testing site to minimize external interference and ensure reproducible measurements.

As shown in Table~\ref{tab:real} and Fig.~\ref{real}, attack effectiveness in the physical domain is consistently lower than in the digital domain across all scenarios. This gap is expected and mainly arises from practical factors in real-world deployment, including printing artifacts and color inaccuracies, placement misalignment, and variations in real-world imaging conditions such as illumination, shadows, and camera exposure. Despite these unavoidable sources of degradation, the proposed method still achieves non-trivial MRSR values in the physical domain. In particular, the pedestrian scenario retains a high MRSR of 0.42, while the cyclist and vehicle scenarios also exhibit consistent depth distortion. These results demonstrate that the adversarial effect is not limited to idealized digital conditions, but can survive the entire physical sensing pipeline.

We also observe that different target categories exhibit varying levels of robustness under physical deployment. Pedestrian-related attacks consistently achieve higher MRSR than cyclist and vehicle scenarios. A plausible explanation is that pedestrians are often less visually prominent in driving scenes and are associated with weaker geometric and semantic priors in monocular depth estimation models. As a result, depth predictions for pedestrians rely more heavily on local appearance cues, making them more susceptible to adversarial perturbations. In contrast, vehicles and cyclists typically occupy larger image regions and exhibit more distinctive structural patterns. These stronger geometric regularities provide implicit constraints for depth inference, which can partially suppress the influence of physically deployed adversarial objects and lead to lower MRSR under physical attacks.

Overall, these results confirm that our method is not only effective in the digital domain, but also physically realizable. Despite inevitable real-world degradations, the proposed attack remains capable of inducing meaningful depth misestimation, posing a tangible risk to real-world autonomous driving systems.

\section{CONCLUSION}

In this work, we introduced a novel training-free framework for challenging the robustness of MDE by formulating adversarial attack as a conditional generative problem rather than patch optimization. Our approach enables the generation of visually coherent adversarial objects at arbitrary locations in the scene, guided jointly by a diffusion prior and the proposed Jacobian Vector Product Guidance, which modulates adversarial influence according to the local score-field geometry. Together with the Salient Region Selection
algorithm, our framework produces substantial depth shifts while preserving strong image realism.

Extensive evaluations on digital domain, along with real-world experiments, demonstrate that the generated adversarial objects induce substantial shifts in the estimated depth across diverse MDE architectures and reliably transfer from the digital to the physical domain. Future work may proceed in two directions. First, the proposed adversarial generation framework can be extended to black-box settings. Second, the generated object-level adversarial scenes can be leveraged as challenging training data to improve the robustness of MDE models, ultimately contributing to the development of safer and more reliable perception systems for intelligent transportation.

\bibliographystyle{IEEEtran}
\bibliography{main}

\end{document}